\documentclass[a4paper, 10pt, conference]{ieeeconf}      

\IEEEoverridecommandlockouts                              

\usepackage{graphics} 
\usepackage{epsfig} 
\usepackage{mathptmx} 
\usepackage{times} 
\usepackage{amsmath} 
\usepackage{amssymb}  
\usepackage{optidef}
\usepackage{xcolor}
\usepackage{algorithm}
\usepackage{algorithmicx}
\usepackage{algpseudocode}
\usepackage{romannum}
\usepackage{verbatim} 
\usepackage{mathtools}
\usepackage{tikz}

\makeatletter
\let\NAT@parse\undefined
\makeatother
\usepackage[colorlinks = false,
            linkcolor = blue,
            urlcolor  = blue,
            citecolor = blue,
            anchorcolor = blue
            ]{hyperref}

\usepackage{algpseudocode}
\usepackage{caption}
\usepackage{multicol, blindtext} 

\newsavebox{\tempbox}

\DeclarePairedDelimiter{\ceil}{\lceil}{\rceil}

\floatname{algorithm}{Algorithm}

\usepackage[noadjust]{cite}

\usepackage{epsfig}
\usepackage{epstopdf}
\DeclareMathOperator*{\argmin}{arg\,min}

\title{\LARGE \bf
A Game Theoretic Approach for Parking Spot Search \\ with Limited Parking Lot Information
}

\author{Yutong Li$^{1}$, Nan Li$^{1}$, H. Eric Tseng$^{2}$, Suzhou Huang$^{2}$, Ilya Kolmanovsky$^{1}$, Anouck Girard$^{1}$, Dimitar Filev$^{2}$
\thanks{*This work is supported by Ford company.}
\thanks{$^{1}$Y. Li, N. Li, I. Kolmanovsky, and A. Girard are with the Department of Aerospace Engineering, University of Michigan, Ann Arbor, MI 48105, USA. Emails:
        \{\tt\small{yutli, nanli, ilya, anouck}\}@umich.edu}%
\thanks{$^{2}$H. E. Tseng, S. Huang and D. Filev are with the Ford Motor Company, Dearborn, MI, 48126 USA. E-mails: \{\tt\small{htseng, shuang10, dfilev}\}@ford.com}
}

\begin{document}

\maketitle
\thispagestyle{empty}
\pagestyle{empty}

\begin{abstract}

We propose a game theoretic approach to address the problem of searching for available parking spots in a parking lot and picking the ``optimal'' one to park. The approach exploits limited information provided by the parking lot, i.e., its layout and the current number of cars in it. Considering the fact that such information is or can be easily made available for many structured parking lots, the proposed approach can be applicable without requiring major updates to existing parking facilities. For large parking lots, a sampling-based strategy is integrated with the proposed approach to overcome the associated computational challenge. The proposed approach is compared against a state-of-the-art heuristic-based parking spot search strategy in the literature through simulation studies and demonstrates its advantage in terms of achieving lower cost function values.

\end{abstract}

\section{Introduction}\label{sec:1}

Parking has been becoming increasingly challenging for drivers with the rapid population/car ownership growth in urban areas. According to a report from INRIX, in the US, a driver spends on average 17 hours a year searching for parking spots \cite{c1}. Also, the limited visibility in a narrow parking space causes high chances of vehicle collisions -- about 40\% of car accidents with physical loss or damage occur during parking \cite{c2}. 

Automated parking systems, which enable vehicles to find available parking spots and park automatically, provide a promising solution. Extensive research has been conducted to address trajectory planning and control problems related to automated parking. For instance, to generate collision-free trajectories in parking lots, heuristic search-based \cite{c3}, sampling-based \cite{c4} or optimization-based approaches \cite{c5} may be adopted. Meanwhile, automobile manufacturers have been equipping new commercial vehicles with various automated parking functionalities, such as the Ford Enhanced Active Park Assist \cite{c6} and the Mercedes Remote Parking Assist \cite{c7} -- both systems enable the vehicle to park into a specified parking spot even when the driver has left the vehicle. In addition to this, there have been several research efforts and advancements in parking guidance systems \cite{c16,c17}, which aim to assist drivers in finding available parking spots more efficiently, typically through acquiring necessary information from the infrastructure, processing it, and displaying the suggested parking spot location as well as the corresponding route to drivers. 

All the above research treats the case where the target parking spot has been assigned in advance. To achieve a fully autonomous parking system, the parking spot search problem needs to be addressed. One solution is to develop centralized managers/coordinators to assign an available parking spot to each vehicle entering the parking lot \cite{c8,c9,c10}. This way, only collision avoidance is handled by the vehicles. However, monitoring the availability of each parking spot and vehicle-to-parking lot communications are needed, requiring major hardware/software updates to existing parking lots. In \cite{c11}, a trajectory coordination strategy for automated parking is proposed, which relies purely on communications among vehicles but not on communications between vehicles and the parking lot. In \cite{c12}, the competition among multiple vehicles for available parking spots is formulated as a game and a Nash equilibrium-based strategy is solved for each vehicle, under the assumption that only limited information of other vehicles' trajectories is known. However, both strategies of \cite{c11} and \cite{c12} assume full knowledge of the locations of available parking spots, which may not be the case in reality due to the limited visibility in a parking lot -- the lines of sight of sensors are likely to be blocked by parked vehicles.

To address the above limitations, we treat in this paper the problem of searching for available parking spots and picking the ``optimal'' one to park under the assumption that only limited information of the parking lot is available. Specifically, we assume that the parking lot layout and the current number of parked cars in it are known, which, as a matter of fact, have been available/can be easily available to many existing parking lots \cite{c13,c14}, but the distribution of parked cars\footnote{i.e., the distribution of available and unavailable parking spots.} in the parking lot is unknown.

Our approach is based on modeling the ego vehicle's problem of parking under uncertainty in the actual distribution of available spots as a game between the vehicle and the parking lot. The ego vehicle's objective is represented using a cost function to minimize, and the parking lot's objective is to cause the highest cost. Through such a game formulation, the effect of worst-case distributions of available spots that may be encountered by the ego vehicle upon each of its parking plans is modeled and compensated in the parking plan selection. Furthermore, the game is repeatedly re-formulated and re-solved to update the parking plan so that the information gathered during the parking process gets fully utilized.

The main contributions of this paper can be summarized as follows: (\romannum{1}) A game theoretic scheme is proposed for solving the problem of parking under limited parking lot information. (\romannum{2}) Within the game theoretic scheme, two different parking strategies are described and compared. (\romannum{3}) A sampling-based method is integrated with these parking strategies to enhance their scalability of treating large parking lots. (\romannum{4}) The advantages of the proposed game theoretic scheme are demonstrated in simulations versus a state-of-the-art heuristic-based parking strategy.

The paper is organized as follows: Section~\ref{sec:2} presents the problem formulation for parking spot search with limited parking lot information. Section~\ref{sec:3} introduces the game theoretic approach to solving the parking problem formulated in Section~\ref{sec:2}. Section~\ref{sec:4} illustrates performance of the proposed approach through simulations. Finally, Section~\ref{sec:5} concludes the paper.


\section{Problem Formulation}\label{sec:2}
\subsection{Parking lot representation}\label{sec:21}

\begin{figure}[thpb]
      \centering
      \includegraphics[scale=0.37]{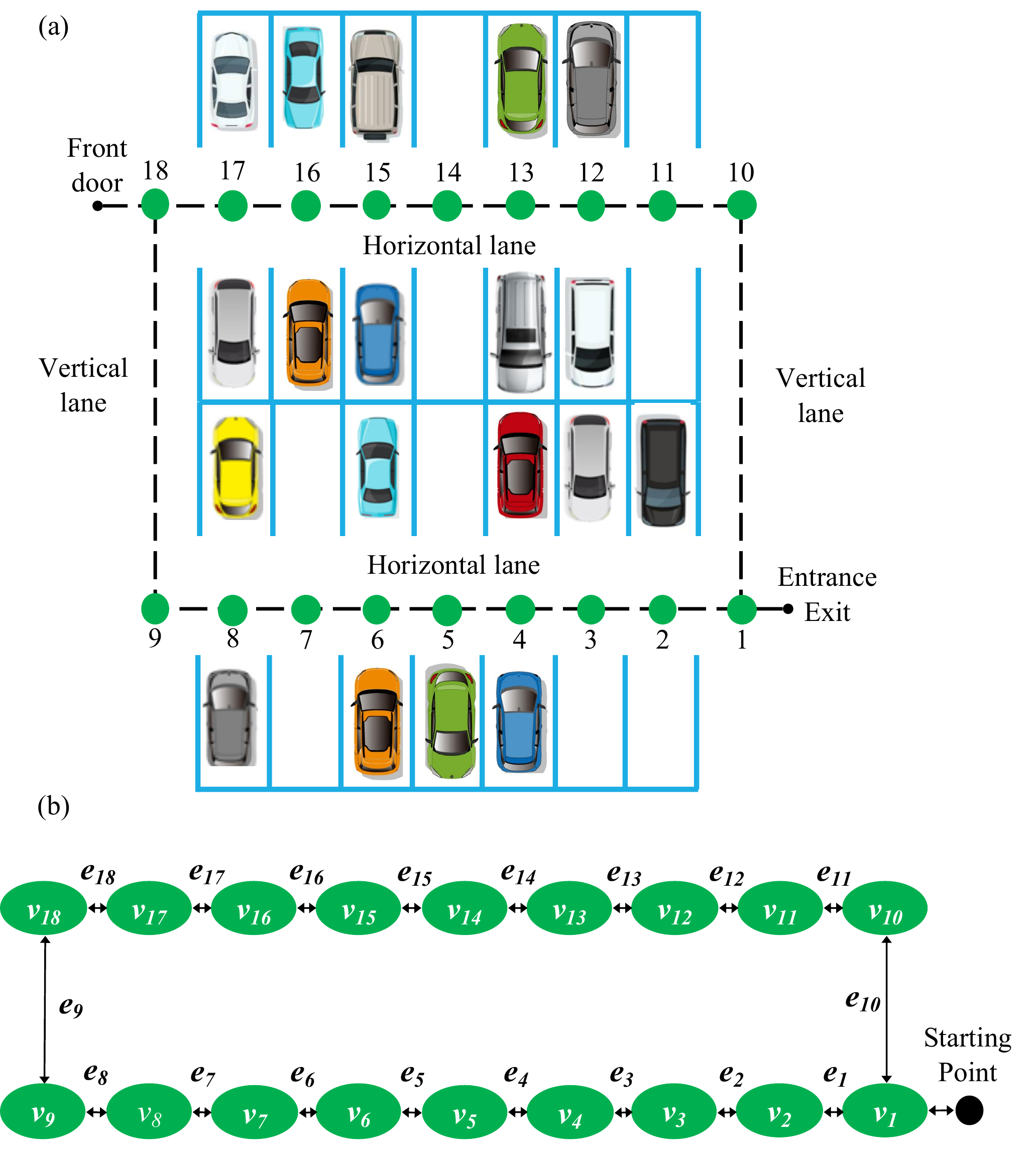}
      \caption{Layout and graph representation of a typical parking lot.}
      \label{fig:parking_lot_graph}
\end{figure}

Before formulating the parking spot search problem, we introduce assumptions regarding the parking lot and the ego vehicle as follows: Firstly, the parking lot has only one entrance/exit and has several horizontal lanes. In particular, both sides of a horizontal lane have parking spots. Secondly, the parking lot knows the total number of parking spots it has and the current number of parked vehicles in it, which is equivalent to knowing the numbers of parking spots that have and have not been occupied (called, respectively, ``unavailable'' and ``available'' spots). However, the distribution of parked vehicles, i.e., whether a specific parking spot is available or occupied, is not known. Note that this is the case for many structured parking lots \cite{c13,c14}. Then, we assume that after entering the parking lot, the ego vehicle chooses a horizontal lane to drive into and search for available spots, and once entering a horizontal lane, the vehicle will either drive into an available spot of that lane and park or drive all the way through that lane, i.e., U-turns in the lane are not allowed \cite{c18}. In particular, the ego vehicle perceives whether a parking spot is available or occupied when it arrives at the front of that spot, provided that this is the first time for the vehicle to arrive at/passing by that spot. Note that although sensors such as radar and camera may have longer perception ranges, the lines of sight to farther parking spots are often blocked by parked vehicles. Therefore, only being able to perceive the availability of nearest spots is a reasonable assumption. After the vehicle reaches the end of a horizontal lane, it chooses another lane to explore. The roads that connect horizontal lanes are called ``vertical lanes.'' A typical parking lot considered in this paper is illustrated in Fig.~\ref{fig:parking_lot_graph}(a).

On the basis of the above assumptions, we model a parking lot as a graph $G = (V,E)$ consisting of a set $V = \{v_1,...,v_{n_V}\}$ of nodes and a set $E = \{e_1,...,e_{n_E}\}$ of edges. A node $v_i \in V$ corresponds to a position in a lane of the parking lot and an edge $e_k = (v_i,v_j) \in E$ represents the immediate connection between the nodes $v_i,v_j \in V$ through lanes. In particular, an edge exists between $v_i$ and $v_j$ when and only when $v_i,v_j$ correspond to positions next to each other in the same lane. Fig.~\ref{fig:parking_lot_graph}(b) illustrates the graph corresponding to the parking lot of Fig.~\ref{fig:parking_lot_graph}(a). Moreover, a binary state $x_i \in \{0,1\}$ is assigned to each node $v_i \in V$ that has associated parking spots to represent its availability, where $x_i = 0$ indicates that the parking spots on both sides of $v_i$ are occupied and $x_i = 1$ means at least one of the two spots is available. For instance, by looking at Fig.~\ref{fig:parking_lot_graph}(a) one can conclude that $x_2 = x_{11} = 1$ while $x_4=0$ in Fig.~\ref{fig:parking_lot_graph}(b). Note that according to the assumptions above, the value of each $x_i$ is not known by the parking lot/ego vehicle.

\subsection{Parking spot evaluation}\label{sec:22}

We are interested in not only finding available parking spots but also picking the ``optimal'' one to park. To evaluate and compare different parking spot options, we consider a cost function to minimize. In particular, the cost function is composed of two parts as follows:
\begin{equation}\label{equ:0}
J(v_{curr},v_{term}) = \omega_r\, J_r(v_{curr},v_{term}) + \omega_t\, J_t(v_{term}),
\end{equation}
where $v_{curr} \in V$ denotes the node corresponding to the ego vehicle's current position, $v_{term} \in V$ denotes the node corresponding to the parking spot option under consideration (called ``target parking spot'') and is the decision variable, $J_r(v_{curr},v_{term})$ is a running cost representing the effort needed by the ego vehicle to get to the target spot, $J_t(v_{term})$ is a terminal cost representing the quality of the target spot, and $\omega_r,\omega_t > 0$ are weighting factors.

In general, how good a parking spot is can depend on various considerations, including individual driver's or parking lot's preferences. As an example, the running cost $J_r$ may correspond to the time and/or fuel spent to drive to the target spot. In this case, $J_r$ can be written as
\begin{equation} \label{eq1}
J_r(v_{curr},v_{term}) = \sum_{k = 1}^{r} c(e_k),
\end{equation}
where $\{e_k\}_{k = 1}^{r}$ is a selected sequence of edges forming a route (or in graph theory terminology, a walk) from the current node $v_{curr}$ to the target node $v_{term}$, and $c(\cdot): E \to \mathbb{R}$ defines the cost associated with each edge, which, for instance, may be determined by the spatial distance between the endpoints of the edge. On the other hand, the terminal cost $J_t$ may correspond to the distance of the target spot to the parking lot entrance/exit, to the entrance door of an office building or a grocery store, or to some facilities such as battery chargers. We remark that our approach can handle any cost function in the form of \eqref{equ:0}, and thus, the cost function can either be pre-designed by the vehicle manufacturer or be specified by individual driver or parking lot.

Note that not all but only the nodes $v_i \in V$ that have available parking spots, i.e., with $x_i = 1$, can be considered as the target node $v_{term}$. Therefore, the problem of picking the optimal parking spot can be cast as the following constrained optimization problem,
\begin{subequations}\label{eq:30}
\begin{align}
\min_{v_i \in V}\,\,\,\,\,\, & \quad J(v_{curr},v_i), \label{eq:3a} \\
\text{subject to} & \quad \,\,\,\, x(v_i) = 1, \label{eq:3c}
\end{align}
\end{subequations}
where $x: V \to \{0,1\}$ defined by $v_i \mapsto x_i$ is to evaluate the binary state value $x_i$ associated with the node $v_i$.

\section{Parking Based on Game against Uncertainty}\label{sec:3}

Corresponding to the assumptions in Section~\ref{sec:21}, the ego vehicle has only limited parking lot information, including the numbers of available and unavailable/occupied parking spots, as well as the availability of each of the spots that the vehicle has arrived at/passed by (called ``visited''). In particular, the availability of each of the parking spots that the ego vehicle has not visited is in general unknown, and as a result, the problem \eqref{eq:30} cannot be solved directly.

Let us use $k \in \mathbb{Z}_{\ge 0}$ to denote decision-execution cycles. Then, the current node of the ego vehicle $v_{curr}$ at the beginning of cycle $k$ can be denoted as $v(k)$. Furthermore, let $V(k)$ denote the set of nodes that have been visited by the ego vehicle and $V \setminus V(k)$ the set of nodes that have not. Let $n_a(k)$ and $n_u(k)$ denote, respectively, the numbers of available and unavailable spots in the unvisited set $V \setminus V(k)$. Based on the assumption that the ego vehicle can perceive the availability of the spots on both sides of a node after it arrives at that node at the end of a cycle, the set $V(k)$ and the values of $n_a(k)$ and $n_u(k)$ can get updated after every cycle. For instance, for the parking lot in Fig.~\ref{fig:parking_lot_graph}, suppose the ego vehicle has visited the nodes $v_1$ to $v_6$, then $V(k) = \{v_1,...,v_6\}$, $V \setminus V(k) = \{v_7,...,v_{18}\}$, $n_a(k) = 6$, and $n_u(k) = 12$. Suppose the vehicle arrives at node $v_7$ after cycle $k$, then $n_a(k+1) = n_a(k)-2$ and $n_u(k+1) = n_u(k)$, since the node $v_7$, which has $2$ available spots, belongs now to the set $V(k+1)$ of visited nodes.

To deal with uncertainty in the distribution of parked vehicles in the unvisited set $V \setminus V(k)$, we exploit a game theoretic approach. It is introduced in detail in what follows.

\subsection{Action spaces of the parking lot and the vehicle}\label{sec:31}

\begin{figure}[thpb]
      \centering
      \includegraphics[scale=0.4]{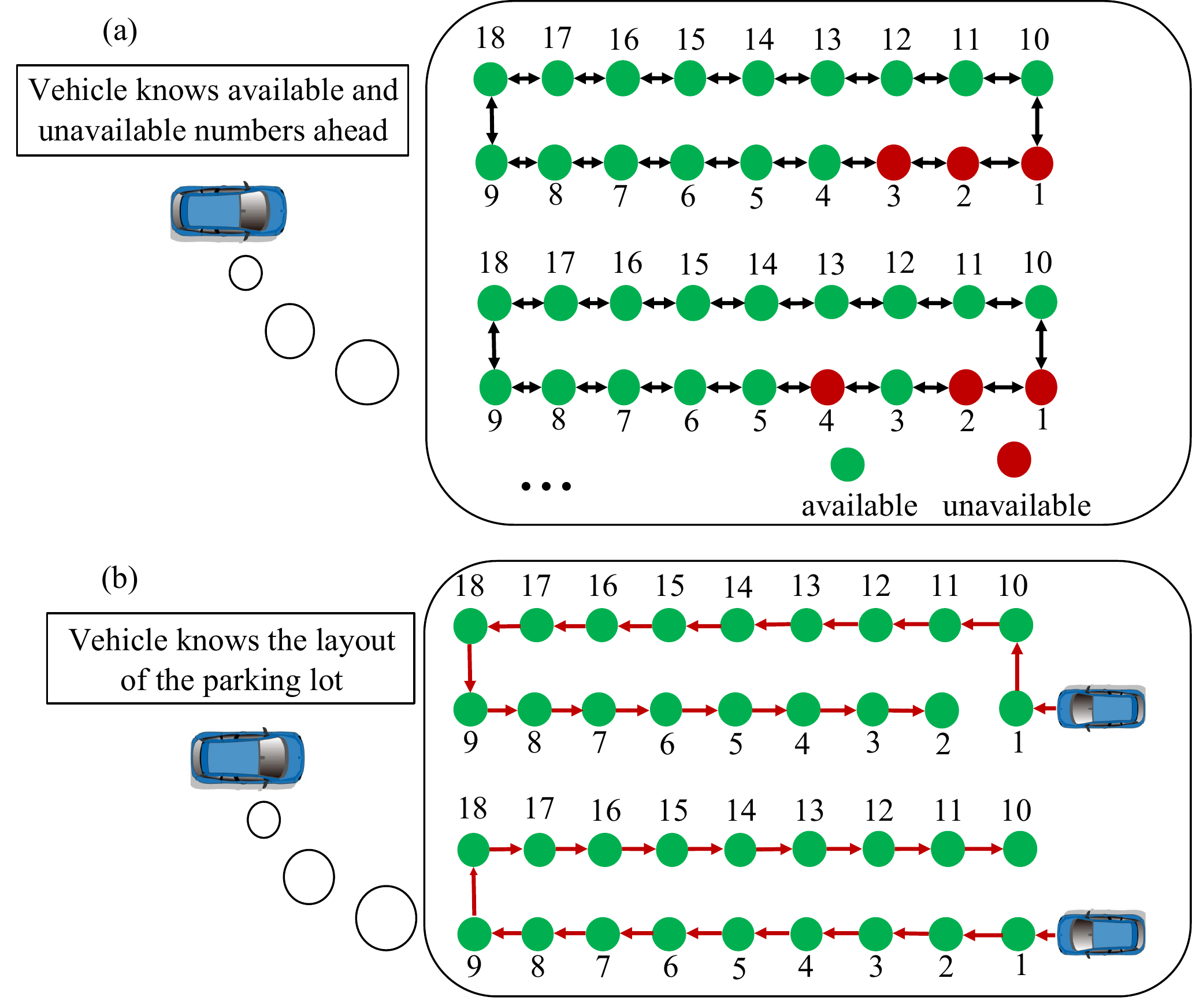}
      \caption{Illustration of parking lot's and ego vehicle's action options. (a) Two possible arrangements for the parking lot of Fig.~\ref{fig:parking_lot_graph} given $n_{x=1}(1) = 15$. (b) Two possible traversal node sequences for the parking lot of Fig.~\ref{fig:parking_lot_graph} at $k = 1$.}
      \label{fig:action_generation}
   \end{figure}

We model the parking lot and the ego vehicle as two agents. For the parking lot, given the numbers of available and unavailable parking spots in the unvisited set $V \setminus V(k)$, $n_a(k)$ and $n_u(k)$, its action space is the set of all possible arrangements. In particular, let $n_{x=1}(k)$ denote the number of nodes in the set $V \setminus V(k)$ with at least one of its two spots available, i.e., with state $x=1$. Given $n_a(k)$ and $n_u(k)$, $n_{x=1}(k)$ can take any value in the set $S_{x=1}(k) = \big\{\ceil[\big]{\frac{n_a(k)}{2}},\ceil[\big]{\frac{n_a(k)}{2}}+1,...,\min(n_a(k),(n_a(k)+n_u(k))/2)\big\}$. Then, an arrangement can be characterized as a way to assign a binary value $x_i \in \{0,1\}$ to each node $v_i \in V \setminus V(k)$ such that $\sum_{v_i \in V \setminus V(k)} x_i = n_{x=1}(k)$. Fig.~\ref{fig:action_generation}(a) illustrates two arrangements for the parking lot in Fig.~\ref{fig:parking_lot_graph}.

For every $n_{x=1}(k) \in S_{x=1}(k)$, the set of all possible arrangements corresponds to the set of all permutations of $n_{x=1}(k)$ ones and $(n_a(k)+n_u(k))/2-n_{x=1}(k)$ zeros. Based on this, we formalize the algorithm for generating the action space $A_p(k)$ of the parking lot as Algorithm~\ref{Algorithm1}. The function \textsc{Permutation}$(n_1,n_0)$ in line 5 generates the set of all componentwise permutations of the vector $(1,...,1,0,...,0)$ with $n_1$ ones and $n_0$ zeros.

\begin{algorithm}
        \caption{Generating the action space of the parking lot}
        \label{Algorithm1}
        \begin{algorithmic}[1] 
            \Require $S_{x=1}(k),n_a(k),n_u(k)$
            \Ensure $A_p(k)$
            \Function {Parking\_lot\_actions}{$S_{x=1}(k),n_a(k),n_u(k)$}
                \State $A_p \gets \emptyset$;
                \For{$n_{x=1}(k) \in S_{x=1}(k)$}
                    \State $n_1\gets n_{x=1}(k)$, $n_0\gets (n_a(k)+n_u(k))/2 - n_1$;
                    \State $A_{temp} \gets \Call{Permutation}{n_1,n_0}$;
                    \State $A_p\gets (A_p,A_{temp})$;       
                \EndFor
                \State \Return{$A_p$}
            \EndFunction
        \end{algorithmic}
\end{algorithm}

We now define the action space of the ego vehicle. In general, the ego vehicle's action space can be defined as the set of all node sequences (starting from its current node $v(k)$) along which the vehicle traverses the graph $G$. To prevent the cardinality of the ego vehicle's action space from growing to infinity, we only consider and include in the set the node sequences that satisfy the following two properties, called ``admissible'': Firstly, the vehicle can only move forward in a horizontal lane, i.e., U-turns in the lane are not allowed; and secondly, the vehicle traverses all nodes without visiting the same horizontal lane twice. For instance, only two node sequences satisfy both properties for the parking lot in Fig.~\ref{fig:parking_lot_graph}, and they are illustrated in Fig.~\ref{fig:action_generation}(b). We remark that although we define an action as a node sequence traversing the graph, the ego vehicle does not necessarily drive along this sequence until the end node: it can choose to stop traversing and park at some node in the middle of the sequence. The reason for defining an action as a complete traversal sequence is to facilitate the subsequent treatment of the problem using game theory.

We leverage the Breadth First Search (BFS) graph traversal algorithm to generate the set $A_v(k)$  of all admissible node sequences for the vehicle. Differently from the standard BFS algorithm, which aims at finding a shortest path from a start node to a goal node, our algorithm for generating the vehicle's action space $A_v(k)$ needs to find all admissible graph traversal paths. In particular, this requires not only the unexplored nodes but also the unexplored horizontal lanes to be stored in the frontiers. Due to space limitation, the algorithm details are omitted but the code can be found at \url{https://github.com/yutlizy/Parking\_Spot\_Search}.

\subsection{Games against uncertainty}\label{sec:32}

Note first that the availability information at the visited set of nodes $V(k)$ is known to the ego vehicle. Then, the set of all possible availability distributions\footnote{i.e., distributions of available and unavailable nodes.} over the unvisited set of nodes $V \setminus V(k)$ has a one-to-one correspondence with the parking lot's actions $a_p \in A_p(k)$. Thus, given a traversal node sequence $a_v \in A_v(k)$, for each $a_p \in A_p(k)$, there is a choice of target node for parking $v_{term} \in V$ among all available nodes that corresponds to the least cost \eqref{equ:0}. Note that since the walk from the vehicle's current node $v(k)$ to each target node option $v_{term}$ is determined by the given node sequence $a_v$, the running cost in the form of \eqref{eq1} associated with the pair $(v(k),v_{term})$ is also uniquely determined. Based on this, we define
\begin{equation}\label{equ:J1}
    \tilde{J}(k,a_v,a_p) = \min_{v_i \in V}\, J\big(v(k),v_i|a_v\big) \text{ subject to } x\big(v_i|a_p,V(k)\big) = 1,
\end{equation}
where $J\big(v(k),v_i|a_v\big)$ is the cost function \eqref{equ:0} with the walk from $v(k)$ to $v_i$ related to the running cost \eqref{eq1} determined by $a_v$, and $x\big(v_i|a_p,V(k)\big)$ is the binary state evaluation function defined in \eqref{eq:3c}, which is determined by $a_p$ as well as the known information associated with the visited set $V(k)$.

If the actual distribution of available and unavailable parking spots is known, which corresponds to a unique $a_p^{actual} \in A_p(k)$, then the optimal action of the ego vehicle can be solved as
\begin{equation}
   a_v^*(k) = \argmin_{a_v \in A_v(k)}\, \tilde{J}(k,a_v,a_p^{actual}),
\end{equation}
and the corresponding optimal parking spot $v^*(k)$ is determined  according to \eqref{equ:J1}.

However, according to our assumptions in Section~\ref{sec:21}, the actual availability distribution, i.e., $a_p^{actual}$, is unknown. In this case, one reasonable strategy for the ego vehicle is to make a decision assuming the worst-case situation. In particular, this can be solved for as
\begin{equation} \label{eq5}
   a_v(k) = \argmin_{a_v \in A_v(k)} \big[\max_{a_p \in A_p(k)}\, \tilde{J}(k,a_v,a_p)\big].
\end{equation}
Since this strategy is to secure that the worst-case performance is optimized, it is named the ``secure strategy.'' 

Once $a_v(k)$ is obtained, the ego vehicle either parks at the current node, in the case where the optimal parking spot corresponding to $a_v(k)$ determined according to \eqref{equ:J1} is the current node $v(k)$ (i.e., the start node of the sequence $a_v(k)$), or drives to the second node in the sequence $a_v(k)$. For the latter, after the new node has been reached, the data $V(k)$, $n_a(k)$ and $n_u(k)$ get updated and the vehicle re-solves \eqref{eq5} with the updated data at the new cycle $k+1$.

The strategy \eqref{eq5} corresponds to a sequential game between the ego vehicle and the parking lot, where after the ego vehicle makes a decision $a_v$, the parking lot makes its own decision $a_p$ to cause the cost $\tilde{J}(k,a_v,a_p)$ as high as possible. Note that this game theoretic model is only used by the ego vehicle to make decisions, the actual availability distribution $a_p^{actual}$ does not change with the game.

\begin{figure}[thpb]
      \centering
      \includegraphics[scale=0.32]{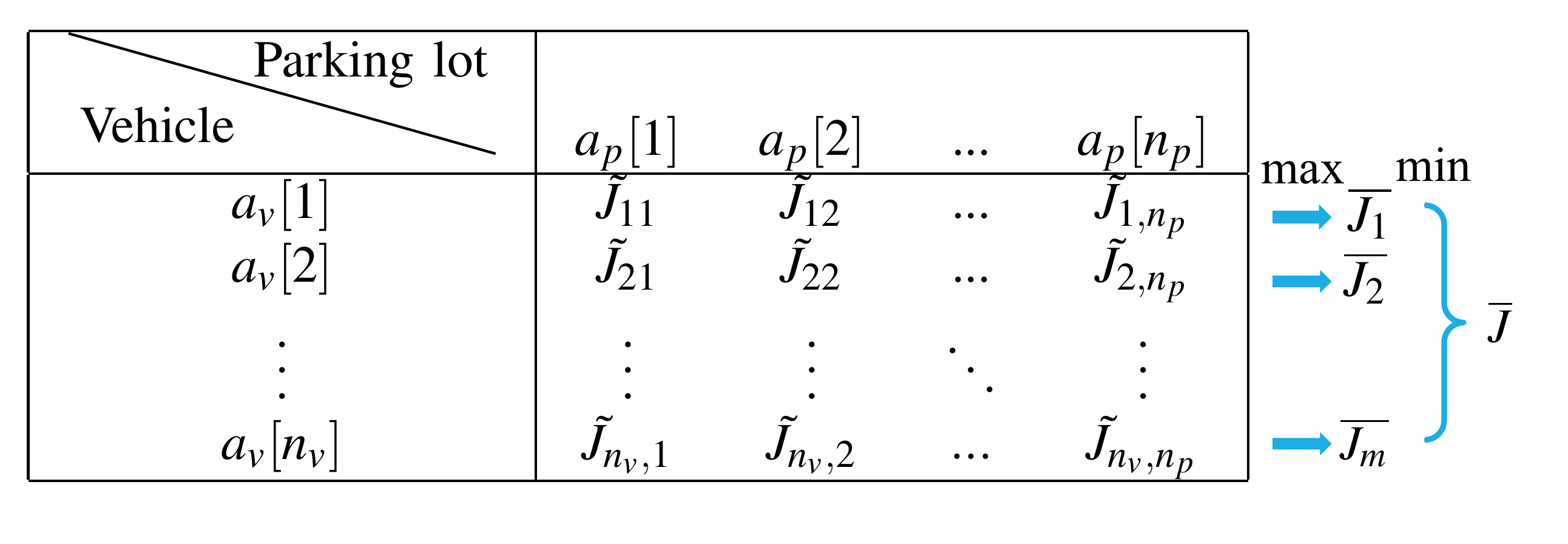}
      \caption{Action selection procedure for the secure strategy.}
      \label{fig:secure}
\end{figure}

The decision strategy \eqref{eq5} can be illustrated by Fig~\ref{fig:secure}.

\subsection{Improved strategy exploiting graph topology of the parking lot}\label{sec:33}

The decision strategy \eqref{eq5} may lead to a conservative solution in terms of rendering a cost higher than what can be possibly obtained. This conservativeness comes from the fact that \eqref{eq5} corresponds to a game where the ego vehicle moves first and the parking lot moves afterwards, and in particular, the parking lot can respond to the ego vehicle's move to cause the highest cost. However, the actual distribution of available and unavailable parking spots is, although unknown to the ego vehicle, fixed and does not change with/respond to the ego vehicle's action. An immediate approach to accounting for this fact and thus mitigating conservativeness is to switch the order of actions of the two players so that the ego vehicle responds to the parking lot's action instead of the opposite. In particular, in the sequel we propose an improvement to the decision strategy, which leads to reduced conservativeness and still maintains the capability of guarding against uncertainty in the availability distribution.

To begin with, note that one important feature of the graph representing the parking lot is that at each node $v(k) \in V$, the ego vehicle has either one or two node options as the next node to travel to (called its travel ``direction''). In particular, when the vehicle is in the middle of a horizontal lane, the only option is the next node in that lane along its travel direction. When the vehicle just exits a horizontal lane, it can choose to go either upward or downward. When the vehicle is traveling along a vertical lane, it can either continue going straight along that vertical lane to the next node if not at the four corners of the parking lot, or turns into a horizontal lane. 

Let us denote the set of directions, i.e., node options as the next node, by $V_{next}(k)$. We gather all actions $a_v \in A_v(k)$ with the same direction $v_{next} \in V_{next}(k)$ as the second node into a group $A_v(k,v_{next})$. Note that the cardinality of $V_{next}(k)$ is either $1$ or $2$, and in turn, there are at most two action groups $A_v(k,v_{next})$.

We let the ego vehicle select its direction according to
\begin{subequations}\label{equ:direc}
\begin{align}
v_{next}(k) &= \argmin_{v_{next} \in V_{next}(k)}\, \overline{J}(v_{next}), \quad \text{where} \label{equ:direc_1} \\
\overline{J}(v_{next}) &= \max_{a_p \in A_p(k)}\, \big[\min_{a_v \in A_v(k,v_{next})}\, \tilde{J}(k,a_v,a_p)\big]. \label{equ:direc_2}
\end{align}
\end{subequations}
Once $v_{next}(k)$ is obtained, the ego vehicle compares $\overline{J}(v_{next}(k))$ determined by \eqref{equ:direc_2} with the cost of parking at the current node $v(k)$ computed from \eqref{equ:0} provided that the current node is available, i.e., $x(v(k)) = 1$. If $\overline{J}(v_{next}(k)) < J(v(k),v(k))$ or $x(v(k)) = 0$, then the ego vehicle proceeds to $v_{next}(k)$; it parks at $v(k)$ otherwise.

The decision strategy \eqref{equ:direc} can be explained as follows: Given a direction $v_{next}$, the parking lot selects its action $a_p \in A_p(k)$ to maximize the vehicle's cost, where for each selection $a_p$, it is assumed that the ego vehicle can optimally respond by selecting $a_v$ in the group $A_v(k,v_{next})$. Note that the maximization over $a_p \in A_p(k)$ accounts for the fact that the actual action of the parking lot, i.e., the actual distribution of available and unavailable parking spots, is unknown -- the ego vehicle estimates the cost of each direction $v_{next} \in V_{next}(k)$ by assuming the worst. Then, on top of this, the ego vehicle selects the direction $v_{next}(k)$ with the lowest worst-case cost.

\begin{figure}[thpb]
      \centering
      \includegraphics[scale=0.32]{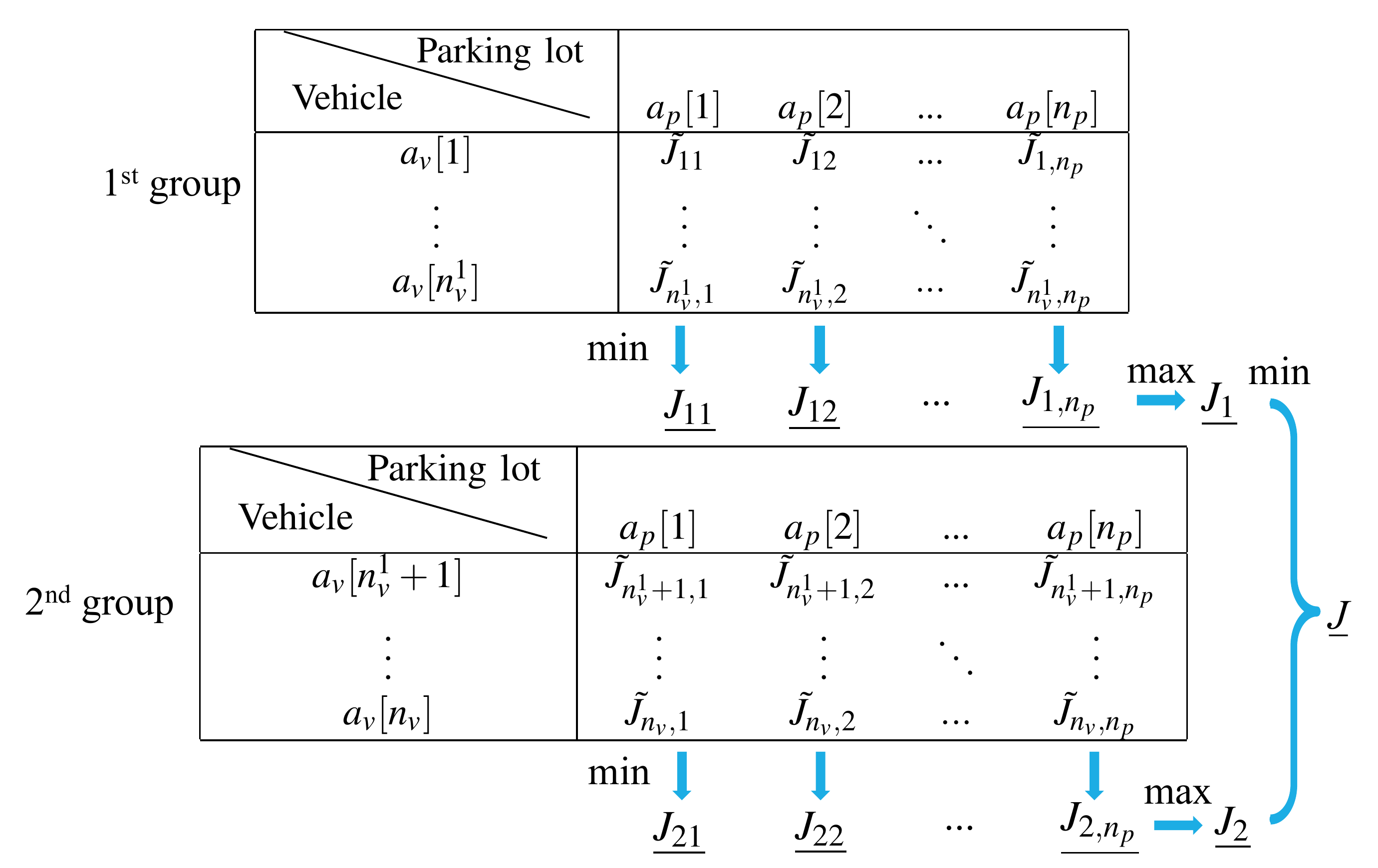}
      \caption{Action selection procedure for the guarded strategy.}
      \label{fig:guarded}
   \end{figure}

The decision strategy \eqref{equ:direc} is further illustrated by Fig~\ref{fig:guarded}. It is named the ``guarded strategy.'' The following Proposition~1 shows that the guarded strategy \eqref{equ:direc} provides less conservative estimates of the costs compared to the secure strategy \eqref{eq5}.

\noindent \textbf{Proposition 1:}
Let $\overline{J}(k)$ and $\underline{J}(k)$ be defined as
\begin{align*}
\overline{J}(k) &= \min_{a_v \in A_v(k)} \big[\max_{a_p \in A_p(k)}\, \tilde{J}(k,a_v,a_p)\big], \\
\underline{J}(k) &= \min_{v_{next} \in V_{next}(k)}\, \max_{a_p \in A_p(k)}\, \big[\min_{a_v \in A_v(k,v_{next})}\, \tilde{J}(k,a_v,a_p)\big].
\end{align*}
Then, $\overline{J}(k) \ge \underline{J}(k)$.

\noindent \textit{Proof:} Note that $\overline{J}(k)$ can be equivalently written as
\begin{equation*}
    \overline{J}(k) = \min_{v_{next} \in V_{next}(k)}\, \min_{a_v \in A_v(k,v_{next})} \big[\max_{a_p \in A_p(k)}\, \tilde{J}(k,a_v,a_p)\big].
\end{equation*}
For each $v_{next} \in V_{next}(k)$, $a_v \in A_v(k,v_{next})$ and $a_p \in A_p(k)$, we have
\begin{equation*}
    \tilde{J}(k,a_v,a_p) \ge \min_{a_v' \in A_v(k,v_{next})}\, \tilde{J}(k,a_v',a_p),
\end{equation*}
and thus,
\begin{equation*}
   \max_{a_p \in A_p(k)}\, \tilde{J}(k,a_v,a_p) \ge \max_{a_p \in A_p(k)}\, \big[\min_{a_v' \in A_v(k,v_{next})}\, \tilde{J}(k,a_v',a_p)\big].
\end{equation*}
Since this holds for every $a_v \in A_v(k,v_{next})$, we have
\begin{align*}
  & \min_{a_v \in A_v(k,v_{next})} \big[\max_{a_p \in A_p(k)}\, \tilde{J}(k,a_v,a_p)\big] \\
  &\quad\quad\quad\quad \ge \max_{a_p \in A_p(k)}\, \big[\min_{a_v' \in A_v(k,v_{next})}\, \tilde{J}(k,a_v',a_p)\big].
\end{align*}
And by the fact that this holds for every $v_{next} \in V_{next}(k)$, we obtain $\overline{J}(k) \ge \underline{J}(k)$. $\blacksquare$

\subsection{Scalability through sampling}\label{sec:34}

For large parking lots with hundreds of parking spots, the action spaces $A_v(k)$, node sequences traversing the graph, and $A_p(k)$, distributions of available and unavailable parking spots, can be large, imposing computational challenge for exactly solving the problem \eqref{eq5} or the problem \eqref{equ:direc}.

To overcome the computational challenge and thus gain scalability to treat large parking lots, we leverage a sampling-based method to approximately solve \eqref{eq5} or \eqref{equ:direc}. In particular, we replace the sets $A_v(k)$ and $A_p(k)$ in the problem formulation with $\tilde{A}_v(k)$ and $\tilde{A}_p(k)$, which are randomly generated subsets of $A_v(k)$ and $A_p(k)$. By restricting the cardinalities of $\tilde{A}_v(k)$ and $\tilde{A}_p(k)$ when randomly generating them, we can bound the computational complexity of the resulting approximated problem.

In particular, as the sizes of $\tilde{A}_v(k)$ and $\tilde{A}_p(k)$ increase and approach the original sets $A_v(k)$ and $A_p(k)$, the solution to the sampled approximation is more likely to agree with the solution to the original problem \eqref{eq5} or \eqref{equ:direc}, with increased computational complexity. 

\section{Simulation Results}\label{sec:4}

We test and evaluate the proposed parking spot search algorithms in simulated parking lots. We first compare the secure and the guarded strategies in terms of their conservativeness. We then illustrate the advantage of our game-theoretic approach over a state-of-the-art heuristic-based parking strategy proposed in \cite{c15}. Finally, we test the feasibility and performance of the sampling-based approach in a large parking lot. 

\subsection{Comparison between secure and guarded strategies}\label{sec:41}

\begin{figure*}[thpb]
\centering
\includegraphics[width=\textwidth]{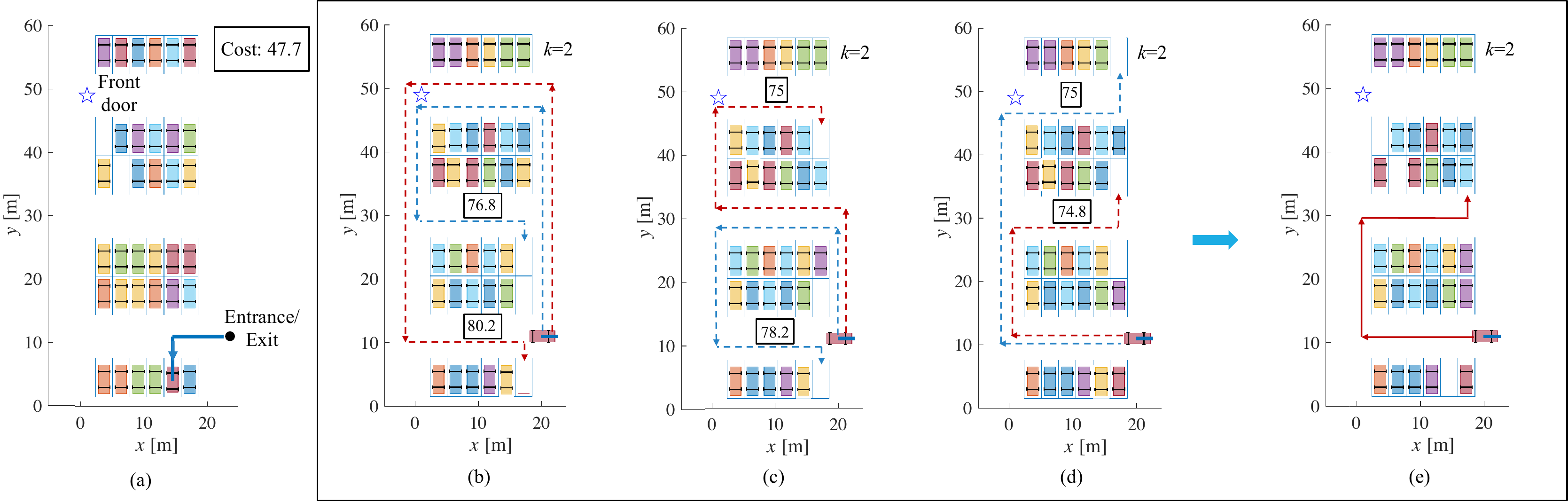}
\caption{Parking result using the secure strategy. (a) The actual distribution of parked vehicles, and the path and parking position of the ego vehicle with a total cost of $44.7$. (b-d) The decision making process at $k=2$. (e) The action selected by the ego vehicle at $k=2$.}
\label{fig:secure_result}
\end{figure*}

We consider a parking lot of a mall shown in Fig.~\ref{fig:secure_result}(a). The actual distribution of parked vehicles is illustrated by the parked cars in the figure. We assume that the front door of the mall is located near the top left corner of the parking lot (marked by the blue star in Fig.~\ref{fig:secure_result}(a)). The terminal cost $J_t$ of each parking spot is assigned as the Euclidean distance between the node associated with the parking spot and the node associated with the front door. Such a terminal cost represents the walking effort of the driver/passenger after parking. The running cost $J_r$ takes the form of \eqref{eq1} where the cost associated with each edge is $1$. Then, the total cost $J$ is computed by \eqref{equ:0}, where $\omega_{r}$ and $\omega_{t}$ are chosen as $1$ and $10$ respectively. 

The parking result using the secure strategy is presented in Fig.~\ref{fig:secure_result}. As shown in Fig.~\ref{fig:secure_result}(a), after entering the parking lot from the entrance, the vehicle drives into the bottom row and parks at the first available spot. The total cost associated with such a parking choice is $47.7$. Figs.~\ref{fig:secure_result}(b)-(d) illustrate the ego vehicle's possible actions $a_v$, i.e., traversal node sequences, and the parking lot's action $a_p$, i.e., a distribution of parked vehicles over the parking lot, that maximizes the cost corresponding to each $a_v$, at $k=2$. In particular, in each of Figs.~\ref{fig:secure_result}(b)-(d), for the two actions $a_v$ represented respectively by the blue and red dashed curves, the corresponding worst-case $a_p$ is the same (illustrated by the parked cars in the figure). Once the corresponding $a_p$ is determined, the ego vehicle identifies the best location $v_i$ associated with the node sequence $a_v$ as its target node for parking (indicated by the endpoints of the dashed curves). After that, the worst-case cost associated with each $a_v$ is computed according to \eqref{equ:J1}, shown in the figures. It turns out that the action corresponding to the red curve in Fig.~\ref{fig:secure_result}(d) has the lowest worst-case cost. Therefore, at $k=2$, the ego vehicle selects that as its action, illustrated by Fig.~\ref{fig:secure_result}(e). Note that although the target node $v_{term}(2)$ associated with the selected action $a_v(2)$ is actually unavailable\footnote{i.e., already occupied by a parked car, as shown in Fig.~\ref{fig:secure_result}(e).} (which is unknown by the ego vehicle at $k=2$), as the vehicle repeatedly updates its information as well as its parking plan by re-solving the problem \eqref{eq5} at every cycle $k$, it finds and parks at an available spot at $k=5$, as shown in Fig.~\ref{fig:secure_result}(a).

\begin{figure*}[thpb]
\centering
\includegraphics[width=\textwidth]{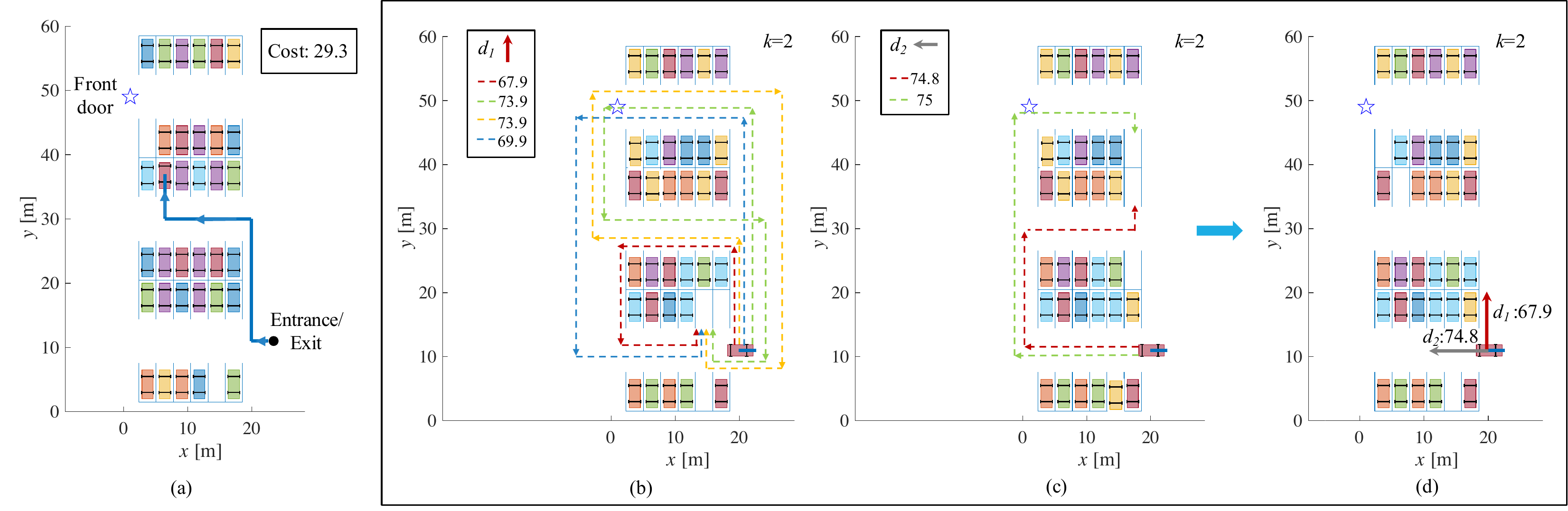}
\caption{Parking result using the guarded strategy. (a) The actual distribution of parked vehicles, and the path and parking position of the ego vehicle with a total cost of $29.3$. (b-c) The decision making process at $k=2$. (d) The action selected by the ego vehicle at $k=2$.}
\label{fig:guarded_result}
\end{figure*}

The parking result using the guarded strategy is presented in Fig.~\ref{fig:guarded_result}. As shown in Fig.~\ref{fig:guarded_result}(a), differently from the result using the secure strategy, the vehicle drives into the middle row and parks there, with a total cost $29.3$. The reason for this choice is illustrated in Fig.~\ref{fig:guarded_result}(b)-(d). At $k=2$, the vehicle has two directions to choose from: going upward or going horizontally forward into the bottom row. For the former, the worst-case action of the parking lot is shown in Fig.~\ref{fig:guarded_result}(b). For such a distribution of parked vehicles, the ego vehicle has four possible actions $a_v$. The path, parking position and corresponding cost associated with each of these four $a_v$ are shown also in Fig.~\ref{fig:guarded_result}(b). It can be seen that the lowest cost is $67.9$. For the direction of going horizontally forward, the worst-case distribution of parked cars, as well as the path, parking position and corresponding cost associated with each of the possible $a_v$ are shown in Fig.~\ref{fig:guarded_result}(c). The lowest cost is $74.8$. Therefore, the ego vehicle chooses to go upward at $k = 2$.

The above simulation results verify that the guarded strategy can find a parking solution with a lower total cost compared to that found by the secure strategy, i.e., the guarded strategy can be less conservative than the secure strategy. This observation also agrees with the theoretical result of Proposition~1. Therefore, from now on, we focus our discussion on the guarded strategy.
    
\subsection{Comparison between guarded and prudent strategies}\label{sec:42}

\begin{figure}[thpb]
      \centering
      \includegraphics[scale=0.73]{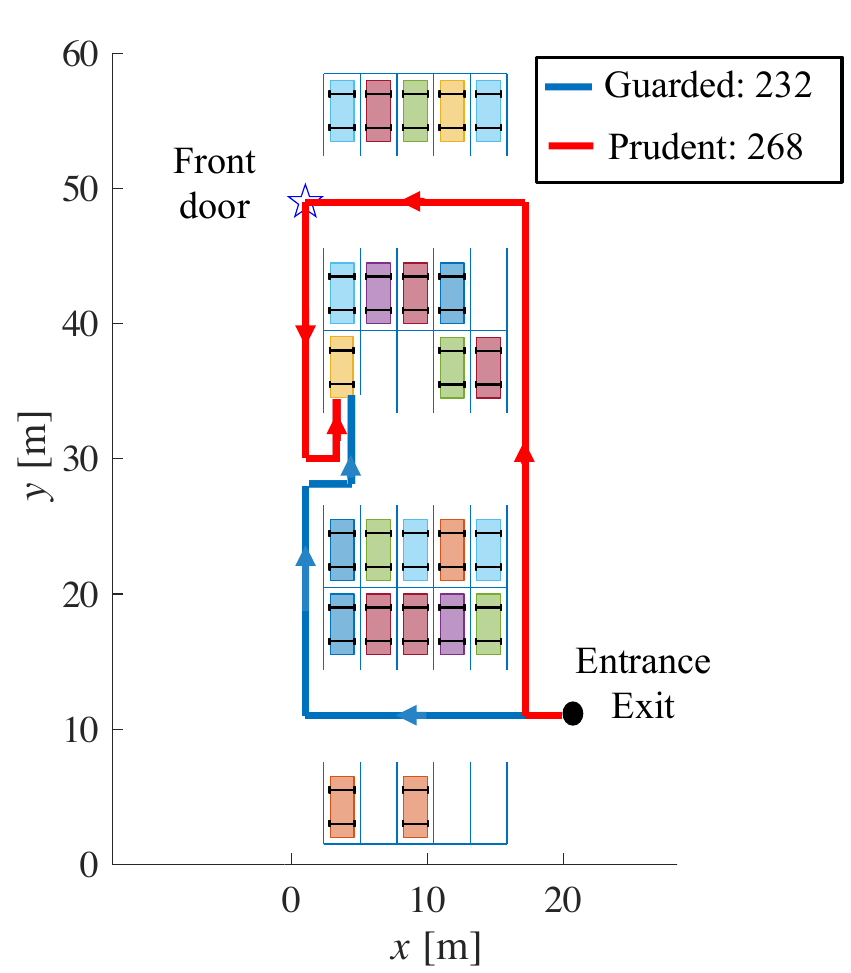}
      \caption{Comparison between the guarded strategy and the prudent strategy.}
      \label{fig:guarded vs prudent}
   \end{figure}

To illustrate the advantage of our proposed game-theoretic approach for parking over heuristic approaches, we compare our approach to the prudent strategy studied in \cite{c15}. The prudent strategy is shown in \cite{c15} to be the optimal among various heuristic parking strategies. With the prudent strategy, the driver starts looking for available parking spots from the row closest to the front door. He always passes the first available spot and hopes that there will be at least one other spot even closer to the front door. If one or more consecutive such available spots are found, he takes the one nearest to the front door to park. Otherwise, the driver backtracks and takes the first available spot he can find. 
The parking lot with the actual distribution of parked cars is shown in Fig.~\ref{fig:guarded vs prudent}. In this study, we assign the cost $c$ associated with each edge $e$ in the running cost \eqref{eq1} as the Euclidean length of that edge, to represent driving effort. We still use the Euclidean distance from the parking node to the front door as the terminal cost of that node, to represent walking effort.  

When the prudent strategy is used, the vehicle drives first into the top row to look for available plots. It skips the first available spot and then finds that there are no other available spots in the top row. So it drives to the second row and parks at the first available spot it meets. 

When the guarded strategy is used, the vehicle parks at the same spot but takes a different path, which results in a lower total cost than that using the prudent strategy (shown at the top right corner of Fig.~\ref{fig:guarded vs prudent}). 

This result shows that our guarded strategy can find a better parking solution than heuristic approaches. Note that our strategy can treat an arbitrary design of cost function in the form of \eqref{equ:0}, which can take into account a wide range of considerations including time and/or fuel consumptions as well as distances to front door and/or facilities. In contrast, the prudent strategy is not as flexible as our approach.

\subsection{Treating large parking lots through sampling}\label{sec:43}

Finally, we show the feasibility of our proposed game-theoretic strategy to treat large parking lots with hundreds of parking spots through the sampling-based method introduced in Section~\ref{sec:34}.

\begin{figure}[thpb]
      \centering
      \includegraphics[scale=0.73]{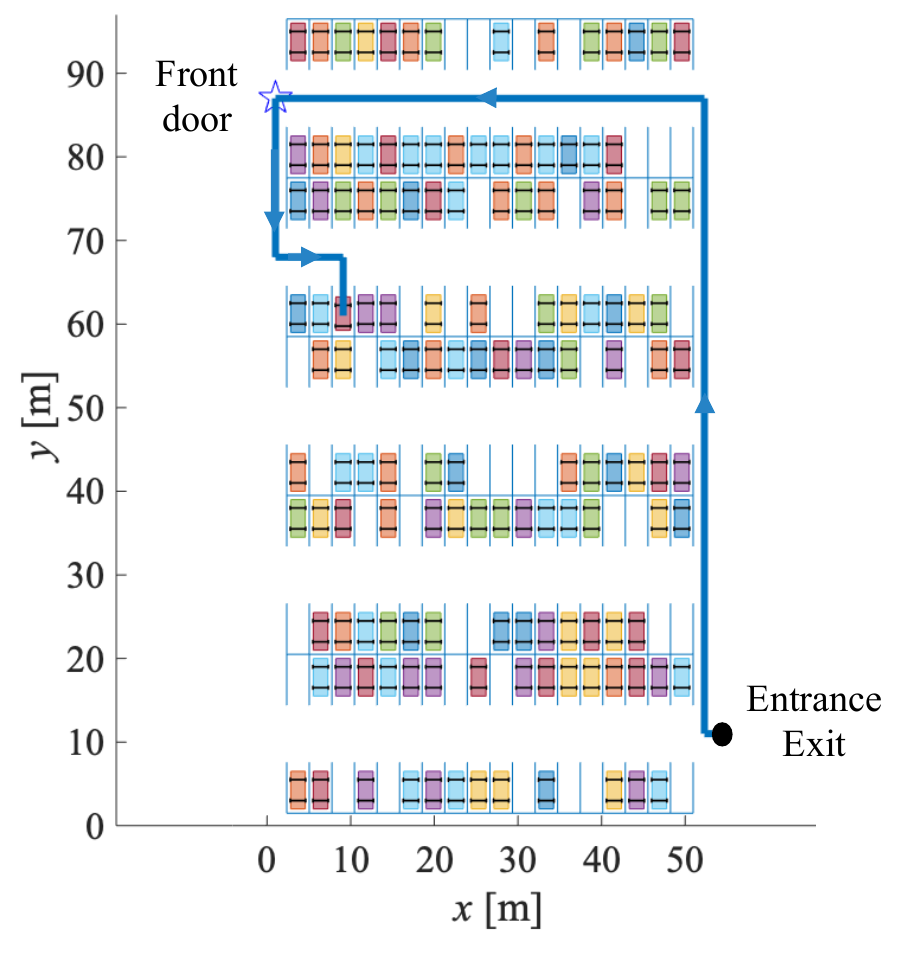}
      \caption{Parking in a large parking lot using the guarded strategy with sampling.}
      \label{fig:sampling}
   \end{figure}

The parking lot under consideration has in total $180$ parking spots and $42$ available ones (see Fig.~\ref{fig:sampling}). At each $k$, we restrict the cardinalities of the sampled subsets $\tilde{A}_v(k)$ and $\tilde{A}_p(k)$ to be at most $1000$. The parking result is shown in Fig.~\ref{fig:sampling}. It can be seen that the vehicle takes a path and parks at a spot, both of which look reasonably good in terms of driving and walking efforts.

\section{Conclusions}\label{sec:5}

In this work, we investigated the problem of parking spot search with limited parking lot information. We treated this problem through a game-theoretic approach, which enabled the vehicle to schedule ``optimal'' parking plans under uncertainty in the distribution of available parking spots. Two strategies, named, respectively, as ``secure'' and ``guarded,'' were proposed and compared. To treat large parking lots, a sampling-based method was adopted to overcome associated computational challenge. Simulation results illustrated good performance of our approach. By comparing against another strategy named ``prudent,'' we also showed that
our systematic, game-theoretic approach could find better parking solutions than heuristic strategies.

\end{document}